\documentclass[runningheads]{llncs}
\usepackage{mathtools}
\usepackage{graphicx}
\usepackage{cite}

\usepackage{url}
\usepackage{graphicx}
\usepackage{pdfpages}
\usepackage{multirow}
\usepackage{booktabs}
\usepackage{amsfonts}
\usepackage[activate={true, nocompatibility}, final, tracking=true,
            kerning=true, spacing=true, factor=1100, stretch=10,
            shrink=10]{microtype}
\usepackage{bm}

\begin{document}
\title{MR Slice Profile Estimation by Learning to Match Internal Patch
Distributions}

\titlerunning{MR Slice Profile Estimation Using Internal Patch
Distributions}

%
%
\author{Shuo Han\inst{1}\orcidID{0000-0003-2033-9004}
\and Samuel Remedios\inst{2}\orcidID{0000-0001-8634-8128}
\and Aaron Carass\inst{3}\orcidID{0000-0003-4939-5085}
\and Michael Sch\"{a}r \inst{4}\orcidID{0000-0002-7044-9941}
\and Jerry L. Prince \inst{3}\orcidID{0000-0002-6553-0876}}

\authorrunning{S. Han et al.}

\institute{Department of Biomedical Engineering, Johns Hopkins
University,\\Baltimore, MD 21218, USA
\and Department of Computer Science, Johns
Hopkins University,\\Baltimore, MD 21218, USA
\and Department of Electrical and Computer Engineering, Johns
Hopkins University,\\Baltimore, MD 21218, USA
\and Department of Radiology, Johns Hopkins University, Baltimore, MD 21205, USA\\
\email{\{shan50, sremedi1, aaron\_carass, mschar3, prince\}@jhu.edu}
}

\maketitle
\begin{abstract}
To super-resolve the through-plane direction of a multi-slice 2D
magnetic resonance~(MR) image, its slice selection profile can be used
as the degeneration model from high resolution~(HR) to low
resolution~(LR) to create paired data when training a supervised
algorithm. Existing super-resolution algorithms make assumptions about
the slice selection profile since it is not readily known for a given
image. In this work, we estimate a slice selection profile given a
specific image by learning to match its internal patch distributions.
Specifically, we assume that after applying the correct slice
selection profile, the image patch distribution along HR in-plane
directions should match the distribution along the LR through-plane
direction. Therefore, we incorporate the estimation of a slice
selection profile as part of learning a generator in a generative
adversarial network~(GAN). In this way, the slice selection profile
can be learned without any external data. Our algorithm was tested
using simulations from isotropic MR images, incorporated in a
through-plane super-resolution algorithm to demonstrate its benefits,
and also used as a tool to measure image resolution. Our code is at
\url{https://github.com/shuohan/espreso2}.
\end{abstract}
\section{Introduction}
To reduce scan time and maintain adequate signal-to-noise ratio,
magnetic resonance~(MR) images of multi-slice 2D acquisitions often
have a lower through-plane resolution than in-plane resolution. This
is particularly the case in clinical applications, where cost and
patient throughput are important considerations. Therefore, there has
recently been increased interest in the use of super-resolution
algorithms as a post-processing step to improve the through-plane
resolution of such images~\cite{xuan-2020-arxiv, zhao-2018-smore,
weigert-2017-fm}. Doing so can improve image visualization as well as
improve subsequent analysis and processing of medical images, i.e.,
registration and segmentation~\cite{zhao-2019-applications}.

Supervised super-resolution algorithms conventionally require a
degeneration model converting high-resolution~(HR) images into
low-resolution~(LR) to help create paired training data. In
multi-slice 2D MR images, this model is often expressed by the slice
profile; indeed, when exciting MR signals using radio frequency~(RF)
pulses within an imaging slice, the slice profile describes the
transverse magnetization of spins along the through-plane
direction~\cite{prince-2006-book}. It acts as a 1D point-spread
function~(PSF) convolved with the object to be imaged, whose full
width at half maximum~(FWHM) is interpreted as the slice thickness.
That is to say, if the slice profile is known, we can then convolve it
with HR images and downsample the resultant volume to match the
through-plane resolution when creating paired training data. Even some
unsupervised super-resolution algorithms~\cite{chen-2020-indirect}
require learning a degeneration model in order to simulate training
data from available images.

Previous methods to estimate the slice profile either require a
simulation using the Bloch equation~\cite{liu-2002-bloch} or
measurement on an MR image of a physical
phantom~\cite{lerski-1989-phantom, college-2018-guide}. When imaging
an object, the effective slice profile can be different from what was
originally designed using the pulse sequence. It has been reported
that a phantom test can allow the slice thickness---represented by the
FWHM of the slice profile---to deviate by as much as 20\% in a
\textit{calibrated} scanner~\cite{college-2018-guide}. Without knowing
the true slice profile, some super-resolution
algorithms~\cite{zhao-2018-smore} assume that it can be approximated
by a 1D Gaussian function whose FWHM is equal to the slice separation.
Xuan~et~al.~\cite{xuan-2020-arxiv}, which takes an alternative
approach, does not take the slice profile into account and reduces the
slice separation without altering the slice thickness.  Given the
potential slice thickness error of 20\% in a calibrated scanner and
even possible slice gaps~(the slice thickness is designed to be
smaller than slice separation) or overlaps~(the slice thickness is
designed to be larger than slice separation), we anticipate that being
able to estimate an image-specific slice profile would help improve
the results of previously proposed super-resolution works. Moreover,
accurate estimation of the slice profile, without any external
knowledge, could be fundamental in calibrating scanners, determining
the exact spatial resolution of acquired MR data, and measuring the
resolution improvement achieved by super-resolution algorithms.

Motivated by previous super-resolution work~\cite{weigert-2017-fm,
zhao-2018-smore}, we note that, in medical imaging, the
texture patterns along different axes of the underlying object can be
assumed to follow the same distributions. In particular, for
multi-slice 2D MR images, this principle indicates that if we use the
correct slice profile, we can degrade patches extracted from the HR
in-plane directions such that the distribution of these patches
will match the distribution of those LR through-plane patches that
are directly extracted from the MR volume. In other words, the slice
profile can be estimated by learning to match patch distributions from
these different directions. Generative adversarial
networks~(GANs)~\cite{goodfellow-2014-gan} have been used to perform
image translation tasks, where they learn to match the distribution of
generated images to the distribution of true images. This motivates us
to use a GAN framework to match internal patch
distributions of an image volume from a multi-slice 2D MR acquisition,
where the slice profile is learned as a part of the generator.

Our work shares a similarity with several recent super-resolution
algorithms. In~\cite{deng-2020-unpaired, chen-2020-indirect}, GANs are
also used to learn the degeneration models. However, they do not
explicitly impose the form of a PSF. Although their algorithms are
capable of learning other image artifacts, these generators are not
guaranteed to obey the physical process of resolution degradation. Our
work is more similar to~\cite{bell-2019-kernelgan}, as they also
explicitly learn a resolution degradation PSF using a GAN.
However, they assume patch recurrence across downsampling scales,
while we base our algorithm upon the similarity along different
directions within the MR image. The contributions of this work are:
\textbf{1)}~we propose an analysis to tie the slice profile estimation
to the matching of the patch distributions in the HR and LR
orientations; \textbf{2)}~we realize this theoretical analysis in
a GAN to learn the slice profile as part of the generator;
\textbf{3)}~we test our algorithm with numerical simulations and
incorporate it into a recent super-resolution algorithm
in~\cite{zhao-2018-smore} to demonstrate its benefits for improving
super-resolution results; \textbf{4)} we show that the proposed
algorithm is capable of measuring the resultant image resolution of
super-resolution algorithms.

\section{Methods}
\subsection{Slice Profile}

\begin{figure}[t]
    \centering
    \includegraphics[width=0.9\textwidth]{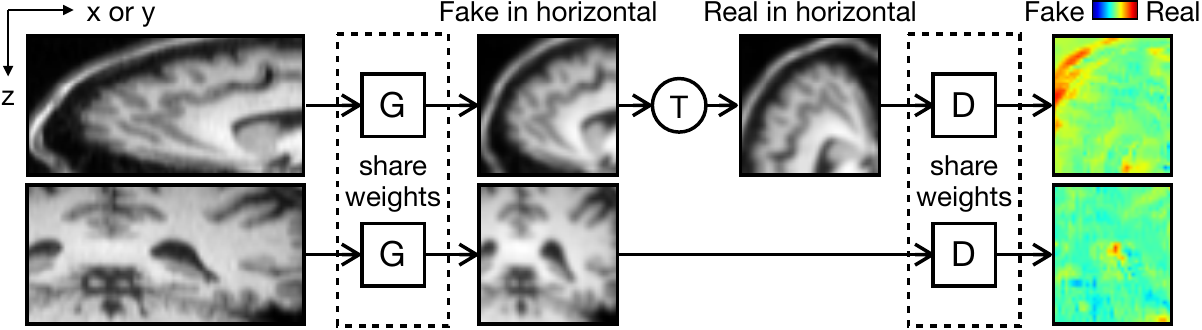}
    \caption{Flowchart of our algorithm. The generator blurs and
    downsamples the \textit{horizontal direction} of an extracted
    image patch, and the discriminator determines whether the
    \textit{horizontal direction} of the input is fake or real LR and
    outputs a ``pixel-wise'' probability map. \textbf{G}: generator;
    \textbf{D}: discriminator; \textbf{T}: transpose.}
    \label{fig:flowchart}
\end{figure}
We regard the 3D object, $f$, imaged in a MR scanner, as being
represented by the continuous function $f(x, y, z)$. We model the
slice selections and signal excitation in a multi-slice 2D acquisition
as a convolution between $f$ and the 1D PSF $p_l(z)$,
\begin{equation*}
f(x, y, z) \ast_z p_l(z).
\end{equation*}
We use $\ast_z$ to denote the 1D convolution along the $z$-axis, since
the symbol $\ast$ generally represents the 3D convolution along $x$-,
$y$-, and $z$-axes. We assume that the truncated k-space sampling
imposes two 1D PSFs $p_h(x)$ and $p_h(y)$, which are identical to each
other but operate on different axes, onto the $xy$-plane,
\begin{equation*}
f(x, y, z) \ast_z p_l(z) \ast_x p_h(x) \ast_y p_h(y),
\end{equation*}
where $\ast_x$ and $\ast_y$ denote the 1D convolutions along the $x$-
and $y$-axes, respectively. In general, the in-plane---the $xy$-plane
in this case---resolution is higher than the through-plane---the
$z$-axis---resolution. This indicates that $p_h$~(usually) has a
smaller FWHM than $p_l$. Additionally, the acquisition of an image
also includes sampling or digitization. In this work, we
represent this as,
\begin{equation}
\left \{ f \ast_z p_l \ast_x p_h \ast_y p_h \right \}
\downarrow_{(s_z, s_x, s_y)},
\label{e:image}
\end{equation}
where $s_x$, $s_y$, and $s_z$ are sampling step sizes along the $x$-,
$y$-,
and $z$-axes, respectively. We further assume that $s_x = s_y < s_z$,
which is the situation in a typical multi-slice 2D acquisition.

\subsection{Slice Profile and Internal Patch Distributions}
\label{sec:patch}

In medical imaging, we observe that $f$ can express similar
patterns~(without any blurring) along the $x$, $y$, and $z$
axes~\cite{zhao-2018-smore, weigert-2017-fm}. We use the following
notations,
\begin{equation*}
I_{xz} = f(x, y, z) \mid_{(x, y, z) \in \Omega_{xz}}
\quad \mathrm{and} \quad
I_{zx} = f(x, y, z) \mid_{(x, y, z) \in \Omega_{zx}},
\end{equation*}
to denote 2D patches extracted from the $xz$- and $zx$-planes,
respectively, where $\Omega_{xz}$ and $\Omega_{zx}$ are the
corresponding coordinate domains of these 2D patches. It should be
self-evident that the patches $I_{xz}$ and their transposes $I_{zx}$
can be assumed to follow the same distribution $\mathcal{F}$. That is,
\begin{equation*}
I_{xz} \sim \mathcal{F}
\quad \mathrm{and} \quad
I_{zx}=I_{xz}^T \sim \mathcal{F}.
\end{equation*}
The self-similar phenomenon implies that the patches extracted from
the $yz$- and $zy$-planes can also be assumed to follow $\mathcal{F}$.
Without loss of generality, we restrict our exposition to the patches
$I_{xz}$ and $I_{zx}$.

In a multi-slice 2D MR image, we cannot directly extract
patches from $f$. Indeed, as noted in Eq.~(\ref{e:image}), the
continuous $f$ is blurred by PSFs then sampled to form a
digital image. Instead, we think of the \textit{sampled} patches as,
\begin{equation}
I_{hl} = \left\{I_{xz} \ast_1 p_h \ast_2 p_l\right\} \downarrow_{(s_x,
s_z)}
\quad \mbox{ and } \quad
I_{lh} = \left\{I_{zx} \ast_1 p_l \ast_2 p_h\right\} \downarrow_{(s_z,
s_x)},
\end{equation}
where $\ast_1$ and $\ast_2$ are 1D convolutions along the first and
second dimensions of these patches, respectively, and
$\downarrow_{(s_x, s_z)}$ represents the sampling along the first and
second dimensions with factors $s_x$ and $s_z$, respectively. This can
be summarized as saying that $I_{lh}$ is blurrier in its first
dimension than its second dimension since $s_z > s_x$, and $p_l$ has
a wider FHWM than $p_h$; in contrast, in $I_{hl}$, the second dimension is
blurrier than its first dimension. Therefore, these two patches,
$I_{lh}$ and $I_{hl}$, cannot be assumed to follow the same
distribution anymore. All is not lost, however. Suppose that we know the
difference between $p_h$ and $p_l$, which we express as another 1D
function $k$, such that $p_h \ast k = p_l$, and the difference between
the sampling rates $s_x$ and $s_z$, defined as $s = s_z / s_x$. In
this case, we can use the self-similar phenomenon and make the
following statements:
\begin{equation}
\left\{I_{hl} \ast_1 k \right\}\downarrow_{(s, 1)} \sim \mathcal{F'}
\qquad \mbox{ and } \qquad
\left\{I_{lh} \ast_2 k \right\}\downarrow_{(1, s)} \sim \mathcal{F'},
\label{e:core}
\end{equation}
where $\mathcal{F'}$ is the resultant patch distribution. This
indicates that after \textit{correctly} blurring then downsampling
each patch, these new patches should follow the same distribution,
namely $\mathcal{F'}$. We note that Eq.~(\ref{e:core}) only holds
approximately; in general, convolution and downsampling
operations do not commute.

Since $k$ is the difference of the through-plane slice profile $p_l$
from the in-plane HR PSF $p_h$, we call $k$ a
``\textit{relative}'' slice profile. We argue that estimating $k$
should be sufficient for the task of improving supervised
super-resolution since we only want to create training pairs from the HR
image or patches that have already been blurred by $p_h$, and $p_h$
may or may not be known. Based on Eq.~(\ref{e:core}), we formulate the
estimation of $k$ as a problem of internal patch distribution
matching. This points us to using a GAN, which is capable of
learning an image distribution during an image translation task.

\subsection{Slice Profile and GAN}

\begin{figure}[t]
    \centering
    \includegraphics[width=0.9\textwidth]{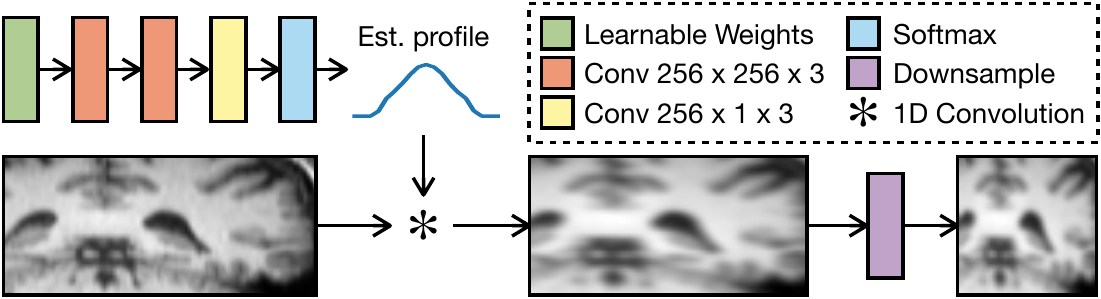}
    \caption{Architecture of our generator. A series of 1D
    convolutions, followed by softmax, outputs the estimated slice
    profile. An image patch is convoled with this slice profile then
    downsampled. \textbf{Conv} $\bm{C_{in} \times C_{out} \times W}$:
    1D convolution along the horizontal dimension with the number of
    input channels $C_{in}$, the number of output channels $C_{out}$,
    and the size of the kernel $W$; \textbf{Est. profile}: the
    estimated slice profile.}
    \label{fig:g}
\end{figure}
According to our analysis in Sec.~\ref{sec:patch}, we can estimate the
\textit{relative} slice profile $k$ by learning to match the internal
patch distributions of a given image volume. In this work, we use a
GAN to facilitate this process. In the following discussion,
we directly use the term ``slice profile'' to refer to $k$ for
simplicity.

The flowchart of our proposed algorithm is shown in
Fig.~\ref{fig:flowchart}. Suppose that the $xy$-axes are HR directions and
the $z$-axis is LR. $xz$ or $yz$ patches are randomly extracted from
the given image volume with sampling probabilities proportional to the
image gradients. As shown in Fig.~\ref{fig:flowchart}, our generator
blurs and downsamples an input image patch only along the horizontal
direction. If the generator output is transposed, its
horizontal direction is the real LR from the image volume; if not
transposed, it means that its horizontal direction is degraded by the
generator. Therefore, we restrict the receptive field of our
discriminator to be 1D, so it can only learn to distinguish the
horizontal direction of the image patch.

\subsubsection{Loss Function.} 
\begin{figure}[t]
    \centering
    \includegraphics[width=0.9\textwidth]{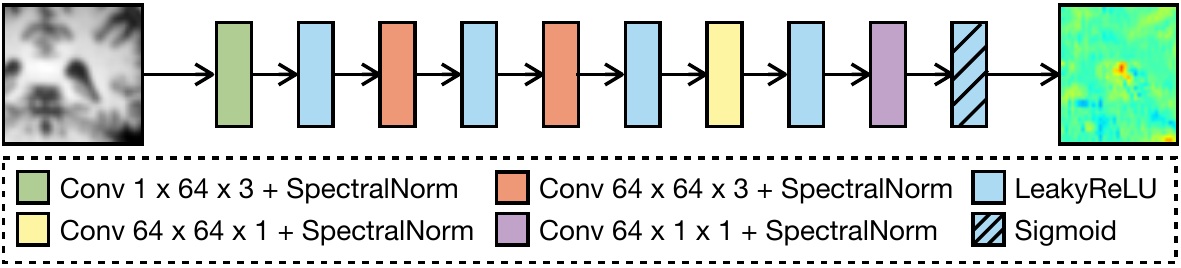}
    \caption{Architecture of our discriminator. This discriminator has
    a \textit{1D horizontal} receptive field and determines whether
    each pixel location \textit{horizontally} corresponds to the real
    LR or is degraded by the generator. \textbf{Conv} $\bm{C_{in}
    \times C_{out} \times W}$: 1D convolution along the horizontal
    dimension with the number of input channels $C_{in}$, the number
    of output channels $C_{out}$, and the size of the kernel $W$;
    \textbf{SpectralNorm}: Spectral normalization.}
    \label{fig:d}
\end{figure}

We updated the conventional GAN value
function of the min-max equation~\cite{goodfellow-2014-gan} to accommodate
our model in Eq.~(\ref{e:core}) as,
\begin{equation}
  \min_G\max_D \left\{ \mathbb{E}_{I}[\log D(G(I)^T)]
              + \mathbb{E}_{I}[\log (1 - D(G(I)))] \right\},
  \label{e:minmax}
\end{equation}
where $G$ is the generator, $D$ is the discriminator, $I$ is a 2D
patch extracted from the image volume, and the superscript $T$
represents transpose. Since the first term in Eq.~(\ref{e:minmax})
includes the generator $G$, we include it in the generator
loss function:
\begin{equation}
  L_{adv} = \log D(G(I_1)^T) + \log (1 - D(G(I_2)))
  \label{e:ladv}
\end{equation}
to learn the generator, where $I_1$ and $I_2$ are two patches
independently sampled from the image. In practice, since our $D$
outputs a pixel-wise probability map~(see Fig.~\ref{fig:d}), we
calculate $L_{adv}$ as the average across all pixels of a patch.

\subsubsection{Generator Architecture.} The architecture of our GAN
generator is shown in Fig.~\ref{fig:g}. Different from previous
methods~\cite{bell-2019-kernelgan, deng-2020-unpaired,
chen-2020-indirect}, we do not take the image patch as the input to
the generator network. Instead, we use the network to learn a 1D slice
profile and convolve it with the patch (without padding). In this way,
we can enforce the model in Eq.~(\ref{e:core}) and impose positivity
and the property of sum-to-one for the slice profile using softmax.
Similar to~\cite{bell-2019-kernelgan}, we do not have non-linearity
between convolutions. Additionally, we want the slice profile to be
smooth. Inspired by the deep image prior~\cite{ulyanov-2018-dip,
cheng-2019-bayesian, ren-2020-selfdeblur}, we note that this
smoothness can be regarded as local correlations within the slice
profile. We then use a learnable tensor~(the green box in
Fig.~\ref{fig:g}) as input to the generator network and simultaneously
learn a series of convolution weights on top of it. In other words, we
can capture the smoothness of the slice profile using only the network
architecture. For this to work, we additionally incorporate an $l^2$
weight decay during the learning of this architecture, as suggested
by~\cite{cheng-2019-bayesian}.

\subsubsection{Discriminator Architecture.} The architecture of the
discriminator is shown in Fig.~\ref{fig:d}. To restrict its receptive
field, we only use 1D convolutions (without padding) along the
horizontal direction. The first three convolutions have kernel size 3,
and the last two have kernel size 1. This is equivalent to a receptive
field of 7, so the discriminator is forced to learn from local
information. To stabilize GAN training, we adjust the convolution
weights using spectral normalization~\cite{miyato-2018-sn} and use
leaky ReLU with a negative slope equal to 0.1 in-between. This
discriminator learns to distinguish whether the horizontal direction
at each pixel is from the real LR or degraded by the generator; it
outputs a ``pixel-wise'' probability map.

\subsection{Regularization Functions and Other Details}

\begin{figure}[t]
    \centering
    \includegraphics[width=0.9\textwidth]{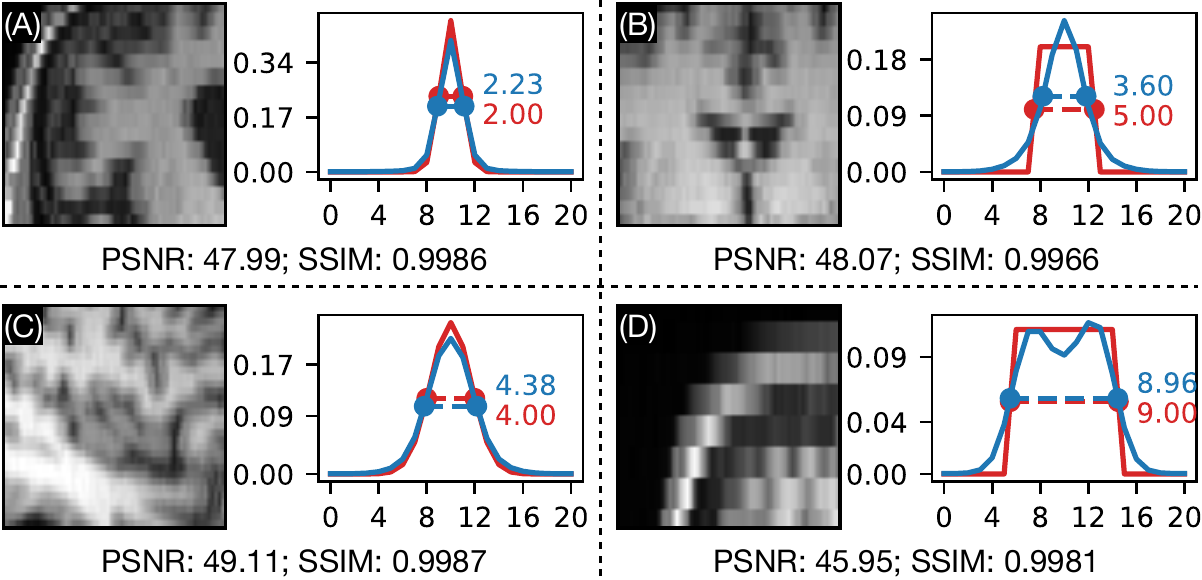}
    \caption{Example estimated slice profiles of the simulations.
    Image slices are shown partially with nearest neighbor upsampling
    for display purposes. True (red) and estimated (blue) slice
    profiles are shown with their corresponding FWHMs. \textbf{(A)}:
    FWHM = 2 mm, scale factor = 4; \textbf{(B)}: FWHM = 5 mm, scale
    factor = 4; \textbf{(C)}: FWHM = 4mm, scale factor = 2;
    \textbf{(D)}: FWHM = 9, scale factor = 8.}
    \label{fig:simu}
\end{figure}

In addition to our adversarial loss in Eq.~(\ref{e:ladv}), we use the
following losses to regularize the slice profile $k$:
\begin{equation*}
  L_c = \left(\sum_{i=0}^{K-1}{ik_i} - \lfloor K  / 2 \rfloor
  \right)^2
  \quad \mathrm{and} \quad
  L_b = k_0 + k_1 + k_{K - 1} + k_{K - 2},
\end{equation*}
where $K$ is the length of $k$ and $\lfloor . \rfloor$ is the floor
operation. $L_c$ encourages the centroid of $k$ to align with the
central coordinate, and $L_b$ encourages its borders to be
zero~\cite{bell-2019-kernelgan}. Note that our $k$ is guaranteed to be
positive and sum to one due to the use of
softmax~(Fig~\ref{fig:g}). Therefore, our total loss for the generator
is
\begin{equation*}
  L = L_{adv} + \lambda_c L_c + \lambda_b L_b + \lambda_w WD,
\end{equation*}
where $WD$ is $l^2$ weight decay, and $\lambda_c = 1$, $\lambda_b =
10$, and $\lambda_w = 0.05$ are the corresponding loss weights,
determined empirically. Inspired by~\cite{yazici-2018-average}, we
further use exponential moving average $\bar{k}$ as our result to
stabilize the estimation,
\begin{equation*}
  \bar{k}^{(t)} = \beta \bar{k}^{(t-1)} + (1 - \beta) k^{(t)},
\end{equation*}
where $t$ is the index of the current iteration, and $\beta = 0.99$. We
use two separate Adam optimizers~\cite{kingma-2017-adam} for the
generator and the discriminator, respectively, with parameters learning rate $= 2
\times 10^{-4}$, $\beta_1=0.5$, and $\beta_2 = 0.999$. Unlike the
generator, the optimizer of the discriminator does not have weight
decay, and the generator optimizer also has gradient clipping. The
number of iterations is 15,000 with a mini-batch size of 64. The size of
an image patch is 16 along the LR direction, and we make sure that
after the generator, the HR direction also has size 16. We initialize
the slice profile to be an impulse function before training.

\begin{table}[t]
\setlength{\tabcolsep}{3.5pt}
\centering
\caption{Errors between true and estimated profiles in the simulations
with Gaussian and rect profiles. The mean values across five subjects
are shown for each metric. \textbf{F. err.}: absolute error between
FWHMs; \textbf{P. err.}: sum of absolute errors between profiles.}
\begin{tabular}{lrrrrrrrrr}
%
\toprule
\multicolumn{10}{c}{\textbf{Gaussian profile}}\\
\midrule
\textbf{FWHM} & \multicolumn{3}{c}{2 mm} & \multicolumn{3}{c}{4 mm}
 & \multicolumn{3}{c}{8 mm} \\
\cmidrule{1-1}
\cmidrule(l){2-4}
\cmidrule(l){5-7}
\cmidrule(l){8-10}
\textbf{Scale} &       2 &       4 &       8 &       2 &       4 &       8 &       2 &       4 &       8 \\
\midrule
\textbf{F. Err}. &    0.32 &    0.41 &    0.69 &    0.36 &    0.26 &    0.52 &    0.97 &    0.23 &    1.33 \\
\textbf{P. Err.} &    0.20 &    0.23 &    0.33 &    0.11 &    0.11 &    0.13 &    0.10 &    0.09 &    0.23 \\
\textbf{PSNR}    &   46.68 &   45.80 &   42.81 &   50.34 &   51.45 &   51.46 &   50.26 &   55.43 &   46.93 \\
\textbf{SSIM}    &  0.9976 &  0.9974 &  0.9957 &  0.9987 &  0.9991 &  0.9993 &  0.9982 &  0.9994 &  0.9975 \\

\midrule
\multicolumn{10}{c}{\textbf{Rect profile}}\\
\midrule

\textbf{FWHM} & \multicolumn{3}{c}{3 mm} & \multicolumn{3}{c}{5 mm}
 & \multicolumn{3}{c}{9 mm} \\
\cmidrule{1-1}
\cmidrule(l){2-4}
\cmidrule(l){5-7}
\cmidrule(l){8-10}
\textbf{Scale} & 2 & 4 & 8 & 2 & 4 & 8 & 2 & 4 & 8 \\
\midrule

\textbf{F. Err} &    0.55 &    0.62 &    0.35 &    0.06 &    1.54 &    1.98 &    0.18 &    0.53 &    0.46 \\
\textbf{P. Err} &    0.44 &    0.44 &    0.47 &    0.28 &    0.41 &    0.45 &    0.25 &    0.24 &    0.23 \\
\textbf{PSNR}   &   45.89 &   46.01 &   43.43 &   47.02 &   46.25 &   45.19 &   46.92 &   50.47 &   48.77 \\
\textbf{SSIM}    &  0.9971 &  0.9976 &  0.9963 &  0.9973 &  0.9972 &  0.9971 &  0.9960 &  0.9987 &  0.9986 \\
\bottomrule

\end{tabular}
\label{tab:simu}
\end{table}

\section{Experiments and Results}

\subsection{Simulations from Isotropic Images}

In the first experiment to test the proposed algorithm, we simulated
LR images from isotropic brain scans of five subjects from the OASIS-3
dataset~\cite{lamontagne-2019-oasis3}. These images are scanned using
the magnetization prepared rapid acquisition gradient echo~(MPRAGE)
sequence with a resolution of 1 mm. We generated two types of 1D
functions, Gaussian and rect, to blur the through-plane direction of
these images. The Gaussian functions have FWHMs of 2, 4, and 8 mm, and
the rect functions have FWHMs of 3, 5, and 9 mm. After blurring, we
downsampled these images with scale factors of 2, 4, and 8. As a
result, we generated $2 \times 3 \times 3 = 18$ types of simulations
for each of the five
subjects. We then ran our algorithm on all these simulated images.
Four metrics were used to evaluate the estimated profiles: the
absolute error between the FWHMs of true and estimated profiles~(FHWM error),
the sum of absolute errors between the true and estimated
profiles~(profile error),
and peak signal-to-noise ratio~(PSNR) and structural
similarity~(SSIM) between the images degraded by the true and
estimated profiles. Note that we calculate SSIM and PSNR within a head
mask. We show these results
in Table~\ref{tab:simu}.  Example images and slice
profiles are shown in Fig.~\ref{fig:simu}.  Despite high FHWM and profile errors,
we observe that the PSNR and SSIM results are very good. We
attribute this disparity to the ill-posed nature of the problem, as
different slice profiles can generate very similar LR images.
\begin{figure}[t]
    \centering
    \includegraphics[width=0.9\textwidth]{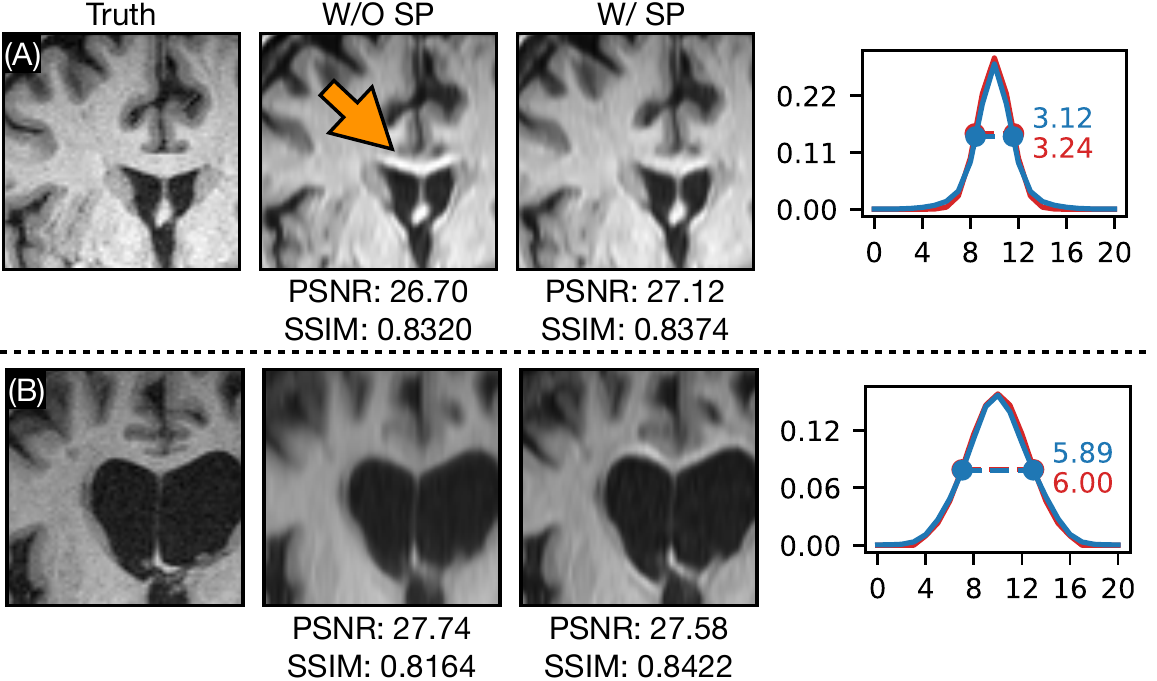}
    \caption{SMORE results. Image slices are shown partially for
    display purposes. True~(red) and estimated (blue) profiles are
    shown with their corresponding FWHMs. The arrow points to an
    artifact. Row \textbf{(A)}: FWHM = 3.2 mm, scale factor = 4; Row
    \textbf{(B)}: FWHM = 6.0 mm, scale factor = 4. \textbf{W/O SP},
    \textbf{W/ SP}: without and with incorporating our estimate
    of the profile into SMORE, respectively.}
    \label{fig:smore}
\end{figure}
\begin{table}[t]
  \centering
  \setlength{\tabcolsep}{4.2pt}
  \caption{SSIM and PSNR between SMORE results and the true HR images
  with and without using the estimated slice profiles. Each metric is
  shown as the mean across five subjects. \textbf{W/ SP}:
  incorporating our estimate of slice profile.}
  \begin{tabular}{lrrrrrrrr}
  \toprule
  \textbf{FWHM} & \multicolumn{2}{c}{2.0 mm} & \multicolumn{2}{c}{3.2 mm} & \multicolumn{2}{c}{4.8 mm} & \multicolumn{2}{c}{6.0 mm} \\

  \cmidrule{1-1}
  \cmidrule(l){2-3}
  \cmidrule(l){4-5}
  \cmidrule(l){6-7}
  \cmidrule(l){8-9}

  \textbf{W/ SP} &   False &   True  &   False &   True  &   False &   True  &   False &   True  \\
  \midrule
  \textbf{PSNR} &   26.11 &   \textbf{27.81} &   27.57 &
  \textbf{27.86} &   \textbf{28.57} &   27.78 &   \textbf{28.15} &
  27.79 \\
  \textbf{SSIM} &  0.8223 &  \textbf{0.8517} &  0.8523 &  \textbf{0.8558} &  \textbf{0.8570} &  0.8565 &  0.8343 &  \textbf{0.8540} \\
  \bottomrule
  \end{tabular}
  \label{tab:smore}
\end{table}

\subsection{Incorporating Slice Profile Estimation Into SMORE}

SMORE~\cite{zhao-2018-smore} is a self-supervised
(without external training data) super-resolution algorithm to improve
the through-plane resolution of multi-slice 2D MR images. It
does not know the slice profile and assumes it is a 1D Gaussian
function with FWHM equal to the slice separation. In this experiment,
we incorporated our algorithm into SMORE and compared the performance
difference with and without our estimated slice profiles. Throughout
this work, we trained SMORE from scratch for 8,000 iterations. We
simulated low through-plane resolution images from our five OASIS-3
scans. We used only Gaussian functions to blur these images with
a scale factor of 4. The Gaussian functions have FWHMs of 2, 3.2, 4.8,
and 6 mm. Therefore, we generated 4 simulations for each of the five
scans.  When not knowing the slice profile, we assume it is a Gaussian
function with FWHM = 4 mm (which is the slice separation in our
simulations). After running SMORE, we calculated the PSNR and SSIM
between the SMORE results and the true HR images as shown in
Table~\ref{tab:smore}. Example SMORE results with and without knowing
the estimated slice profiles are shown in Fig.~\ref{fig:smore}.
Table~\ref{tab:smore} shows better PSNR and SSIM, if our estimated
profiles are used, for simulations with FWHMs of 2 and 3.2 mm. The
PSNR are worse for 4.8 and 6.0 mm, but SSIM is better for FWHM 6.0 mm.

\subsection{Measuring Through-Plane Resolution After Applying SMORE}

\begin{figure}[t]
    \centering
    \includegraphics[width=0.9\textwidth]{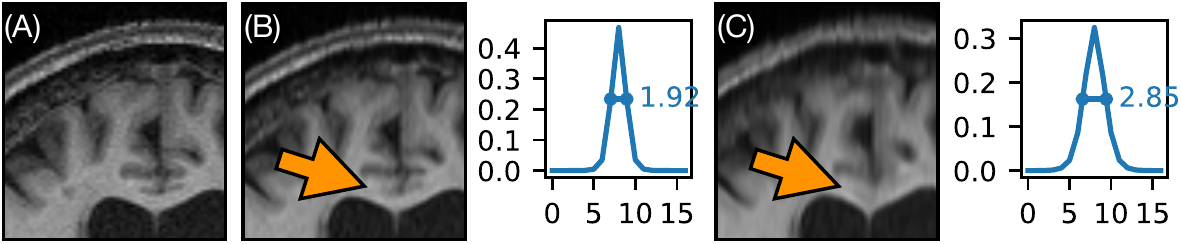}
    \caption{Resolution measurements of super-resolved images. SMORE
    results and the corresponding slice profile estimates from the
    \textit{resultant} images of SMORE are shown in (B) and (C). The
    FWHMs of estimated profiles are marked on the side. The arrows
    point to the differences. \textbf{(A)}: truth HR image;
    \textbf{(B)}: SMORE result from a simulation with FWHM = 2 mm;
    \textbf{(C)}: SMORE result from a simulation with FWHM = 4 mm.}
    \label{fig:measure}
\end{figure}

In this experiment, we use the FWHMs of our estimated slice profiles
to measure the resultant resolution after applying SMORE to an image.
We used Gaussian functions to blur the five OASIS-3 scans. The first
type (Type 1) of simulations has a FWHM of 2 mm and a scale factor of
2, while the second type (Type 2) has FWHM of 4 mm and a slice factor
of 4. We first
used SMORE to super-resolve these images then applied our slice
profile estimation to measure resultant through-plane resolution. Type
1 has mean FWHM = 1.9273 mm with standard deviation (SD) = 0.0583 mm
across the five subjects. Type 2 has mean FWHM = 2.9044 mm with SD =
0.1553 mm. Example SMORE results and resolution measurements are shown
in Fig.~\ref{fig:measure}.

\section{Discussion and Conclusions}

In this work, we proposed to estimate slice profiles of multi-slice 2D
MR images by learning to match the internal patch distributions.
Specifically, we argue that if an HR in-plane direction is degraded by
the correct slice profile, the distribution of patches extracted from
this direction should match the distribution of patches extracted from
the LR through-plane direction. We then proposed to use a GAN to learn
the slice profile as a part of the generator. Our algorithm is
validated using numerical simulations and incorporated into SMORE to
improve its super-resolution results. We further show that our
algorithm is also capable of measuring through-plane resolution.

The first limitation of our algorithm is that it is unable to learn a
slice profile with a flexible shape as shown in Fig.~\ref{fig:simu},
where the true profiles are rect functions. This is because we used a
similar generator architecture as the deep image prior to encourage
smoothness. We found that the value of $\lambda_w$ for weight decay
greatly affected the performance. Specifically, with a small
$\lambda_w$, the shape of the learned slice profile is more flexible, but
the training is very unstable and can even diverge. A better way to
regularize the slice profile should be investigated in the future.

The second limitation is our inconclusive improvement to the SMORE
algorithm. Indeed, we have better PSNR and SSIM when the slice
thickness is smaller than the slice separation~(resulting in slice
gaps when FWHMs are 2 and 3.2 mm in Table~\ref{tab:smore}); this does
not seem to be the case when the slice thickness is larger~(resulting
in slice overlaps when FWHMs are 4.8 and 6.0 mm in
Table~\ref{tab:smore}). We regard this as a counter-intuitive result
and plan to conduct more experiments with SMORE, such as train with
more iterations, and testing other super-resolution algorithms.

\section{Acknowledgments}
This work was supported by a 2019 Johns Hopkins Discovery Award and
NMSS Grant RG-1907-34570.

\bibliographystyle{splncs04}
\bibliography{2021_ipmi_shan_slice-profile}

\end{document}